\documentclass[10pt,twocolumn,letterpaper]{article}

\usepackage[pagenumbers]{cvpr} %

\usepackage{graphicx}
\usepackage{amsmath}
\usepackage{amssymb}
\usepackage{booktabs}
\usepackage{array}
\usepackage{diagbox}

\usepackage{times}
\usepackage{epsfig}
\usepackage{graphicx,subcaption}
\usepackage{diagbox}
\usepackage{multirow}
\usepackage{amsmath}
\usepackage{amssymb}
\usepackage{booktabs}
\usepackage[table,xcdraw]{xcolor}
\usepackage[T1]{fontenc}

\usepackage[pagebackref,breaklinks,colorlinks]{hyperref}
\usepackage{xcolor}
\hypersetup{
    colorlinks,
    linkcolor={red!70!black},
    citecolor={blue!70!black},
    urlcolor={blue!70!black}
}

\usepackage[capitalize]{cleveref}
\crefname{section}{Sec.}{Secs.}
\Crefname{section}{Section}{Sections}
\Crefname{table}{Table}{Tables}
\crefname{table}{Tab.}{Tabs.}

\def\cvprPaperID{10358} %
\def\confName{CVPR}
\def\confYear{2024}

\begin{document}

\title{CRAFT: Contextual Re-Activation of Filters for face recogntion Training}

\author{Aman Bhatta$^1$ \quad\quad Domingo Mery$^2$ \quad\quad Haiyu Wu$^1$  \quad\quad Kevin W. Bowyer$^1$$^{\thanks{Dr. Bowyer is a member of the FaceTec (\url{facetec.com}) Advisory Board.  Results in this paper do not necessarily relate to FaceTec products.}}$ \\\\
$^1$University of Notre Dame\\$^2$Pontificia Universidad Católica de Chile \\
{{\tt\small \{abhatta,hwu6,kwb\}@nd.edu, domingo.mery@uc.cl}}}

\maketitle

\begin{abstract}

The first layer of a deep CNN backbone applies filters to an image to extract the basic features available to later layers. 
During training, some filters may go inactive, meaning all weights in the filter approach zero. An inactive filter in the final model represents a missed opportunity to extract a useful feature. This phenomenon is especially prevalent in specialized CNNs  such as for face recognition (as opposed to, e.g., ImageNet). For example, in one the most widely face recognition model (ArcFace), about half of the convolution filters in the first layer are inactive. We propose a novel approach designed and tested specifically for face recognition networks, known as ``CRAFT: Contextual Re-Activation of Filters for Face Recognition Training''. CRAFT identifies inactive filters during training and reinitializes them based on the context of strong filters at that stage in training.  We show that CRAFT reduces fraction of inactive filters from 44\% to 32\% on average and discovers filter patterns not found by standard training. Compared to standard training without reactivation, CRAFT demonstrates enhanced model accuracy on standard face-recognition benchmark datasets including AgeDB-30, CPLFW, LFW, CALFW, and CFP-FP, as well as on more challenging datasets like IJBB and IJBC.

\end{abstract}

\section{Introduction}
\begin{figure}[t]
    \centering
    \includegraphics[width=0.95\columnwidth]{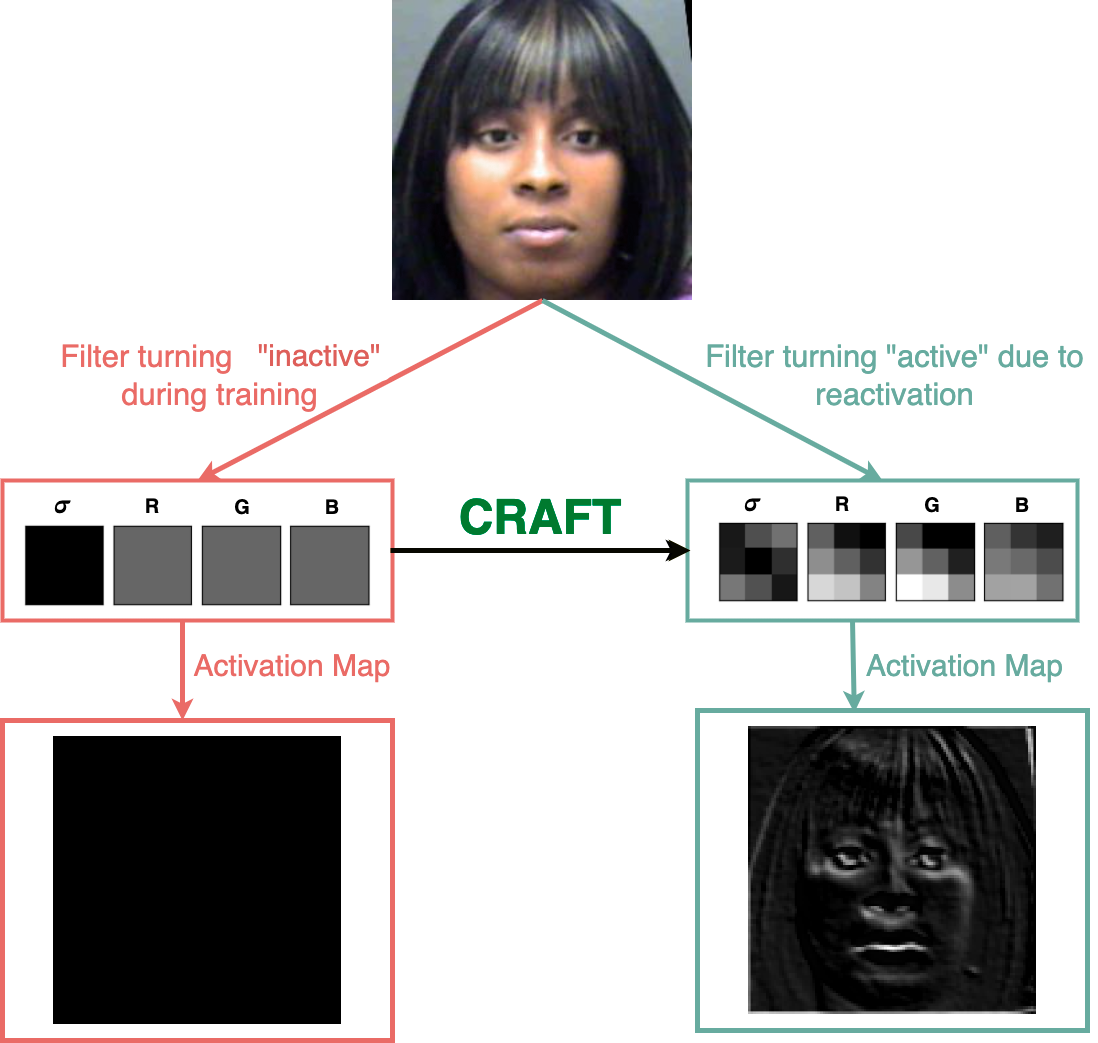}
    \caption{{\bf CRAFT: Contextual Re-Activation of Filters for face recognition Training.}
Face image at top is  example input to ArcFace model. 
Middle left depicts filter 35 of 64 in the first layer, after epoch 3, detected as  inactive.
The black $3 \times 3$ grid indicates no variation of weights across R, G and B planes of filter, and the three gray $3 \times 3$ grids indicate near-zero values of all weights in each plane.
Once a filter goes inactive, it stays inactive unless reactivated.
Bottom left is the activation map resulting from applying the inactive filter to the top image.
Middle right depicts the final model’s version of this filter after reactivation using our approach. 
Bottom right is the activation map from the reactivated filter.
Our approach learns better models by detecting inactive filters during training and reactivating them to give a second chance to learn a useful filter.
}
    \label{fig:intro_fig}
    \vspace{-0.5em}
\end{figure}

{\it Can we improve deep CNN face training by detecting when a filter in the first layer goes inactive, and reactivating it in the context of other filters in that epoch, as illustrated in Fig. \ref{fig:intro_fig},  so that the final model has fewer inactive filters and achieves greater accuracy?}

The first layer of any deep CNN backbone has weights organized into $N$ filters.
For example, the ResNet backbone in ArcFace \cite{deng_cvpr_2019} has 64 $3 \times 3 \times 3$ filters.
First-layer filters are important because they define the primitive features extracted from an image.
There are implicit assumptions in training a deep CNN that (a) no more than $N$ distinct primitive features are needed to compute a good solution to the problem, and (b)  training will result in learning the appropriate filter weights to compute the features.

\begin{figure}[t]
    \centering
    \includegraphics[width=1\columnwidth]{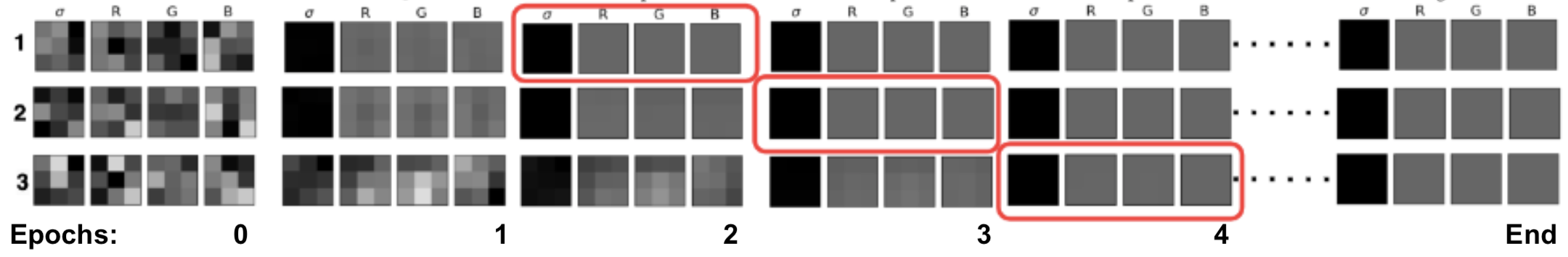}
    \caption{{\bf Illustration of Three Filter Weights Through Training Process.} The red bounding box indicates the epoch where a $3 \times 3 \times 3$ filter goes ``inactive". In standard training,  once the weights in a filter go ``inactive",  the filter is stuck in that state.}
    \label{fig:weight_track}
    \vspace{-0.5em}
\end{figure}

Traditional training often results in a model that has some inactive first-layer filters, in that all the filter weights are near zero and so the filter does not compute any useful feature.
A filter going inactive and being stuck in that state is an effect of using ReLU activation, as discussed in \cite{he_iccv_2015}.
In conventional training, when the set of weights for a filter reaches zero, the filter remains in that state for the duration of the training. Figure \ref{fig:weight_track} shows examples of the $3 \times 3 \times 3$ filter weights during training, from initialization to end of training. 
It is possible to detect filters that go inactive and to reinitialize them to reactivate the filter’s learning.
Based on our experience, reinitializing the weights in the same manner as assigning random weights at the beginning of training is not effective. 
Instead, it is important to select reinitializing weights based on the context of the other, strong filters at that point in training.
This paper shows that our \textit{CRAFT} approach to reactivation produces
face recognition models with fewer inactive filters and that achieve higher accuracy.

Standard CNN training randomly initializes weights across the network with values from a Gaussian distribution with $\mu=0$  and $\sigma=0.1$ \cite{he_iccv_2015}, and updates weights by backpropagation throughout training. A ``Reinitialization’’ of a weight occurs if the value  computed from backpropagation is replaced by a different value.
Our use of reinitialization is fundamentally different in four respects. First, our approach reinitializes a set of weights that represents an entire filter kernel in the first convolution layer, rather than an individual weight from anywhere in the network.  This is important because all weights in filter kernel taken together is what defines a feature extracted by the filter in the first layer.  Second, our reinitialization is triggered by the event of a filter going inactive, rather than occurring at random.  This is important because filters that go inactive represent a missed opportunity to provide a useful feature to later layers of the network.  Third, our reinitialization uses values derived from the context of other strong first-layer filters at that point in training. This is important because the reinitialized filters can then compete on an equal basis with the other strong filters at this point in the learning process.  Finally, our approach is aimed at enabling the training to discover first-layer filters that it would otherwise miss, and so increase accuracy, rather than adding another form of regularization to the training.

In the case of the ResNet backbone used in face recognition networks, our approach reinitializes the $3 \times 3 \times 3 = 27$ weight values of a first-layer filter after the epoch in which all the values have gone below a threshold.
And our approach uses knowledge of the stronger filters in that epoch to reinitialize an inactive filter.
Contributions of this work include:
\begin{itemize}
\item
This is the first approach we are aware of to automatically detect when a first-layer filter in a deep CNN %
goes inactive during training and reinitialize it for another chance to learn a useful feature.
\item
Our  results show that reinitializing inactive filters %
based on the context of active filters in that epoch is an effective strategy, resulting in a model with a lower fraction of inactive filters, discovering additional unique filter patterns, and achieving higher accuracy on benchmark test sets like LFW, CFP-FP, AGEDB-30, CALFW, CPLFW, and on IJB-B and IJB-C. 
\end{itemize}
\begin{figure}[t]
    \centering
    \includegraphics[width=0.9\columnwidth]{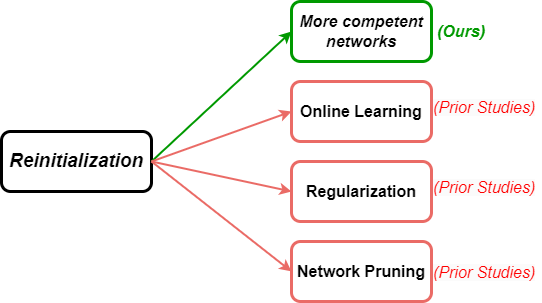}
    \caption{{\bf Existing usage of ``reinitialization in literature vs Ours?"}. Reinitialization finds its primary application in the realm of online learning, sparse networks, and regularization, as commonly portrayed in current literature. However, our approach departs from this convention, as we view reinitialization from a distinct perspective. We utilize the network's current state to facilitate the training of more competent networks.}
    \label{fig:diff_use}
    \vspace{-0.25em}
\end{figure}

This paper is organized as follows. Section \ref{litreview} gives a brief literature review. Section \ref{inactive_in_pretrained} presents the evaluation of the presence of the inactive filter in the pre-trained models. Section \ref{approaches} evaluates multiple approaches to reactivate the inactive filters.  Section \ref{implementation} describes the implementation details of the network use. 
Section \ref{results} - \ref{conclusions} presents the results, analysis and discussion of the results from this work. 

\begin{figure*}[!ht]
 \centering
  \begin{subfigure}[b]{0.9\linewidth}
      \begin{subfigure}[b]{0.32\linewidth}
        \centering
          \includegraphics[width=1\linewidth]{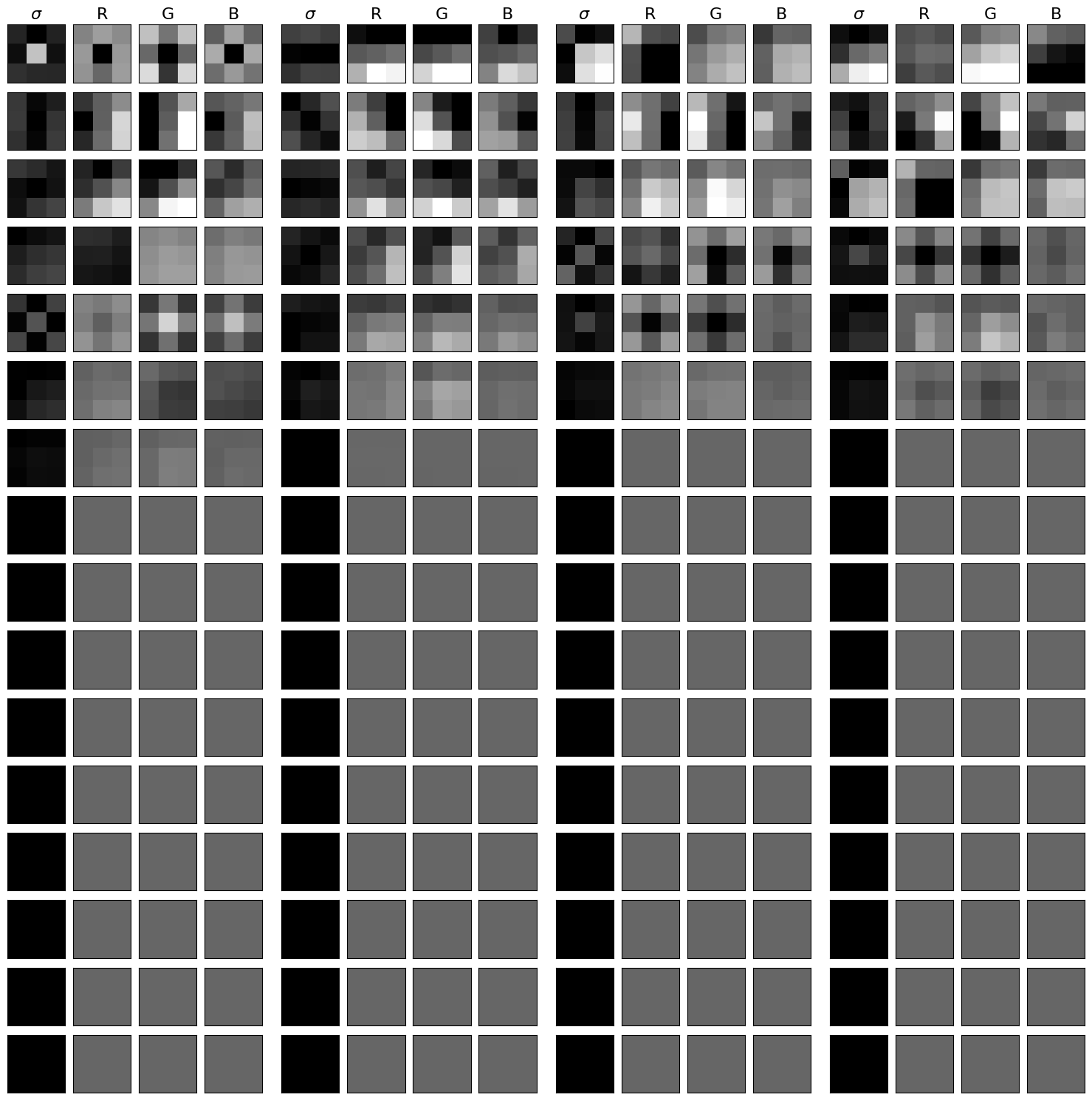}
          \caption{\scriptsize{arcface-r100-ms1mv2}}\label{arcface}
      \end{subfigure}
      \hfill
      \begin{subfigure}[b]{0.32\linewidth}
        \centering
          \includegraphics[width=1\linewidth]{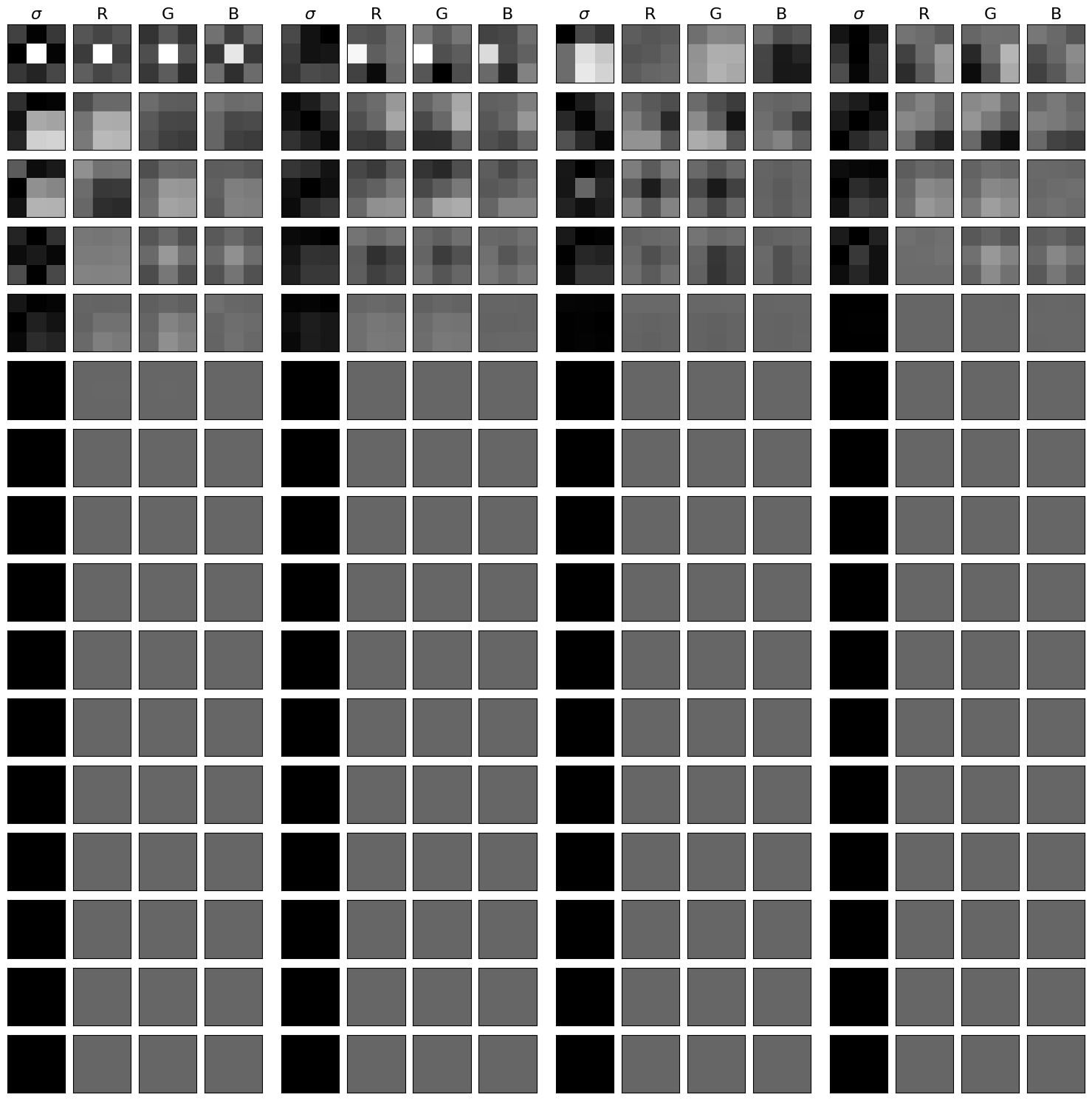}
          \caption{\scriptsize{arcface-r100-ms1mv3}}\label{magface}
      \end{subfigure}
      \hfill
      \begin{subfigure}[b]{0.32\linewidth}
        \centering
          \includegraphics[width=1\linewidth]{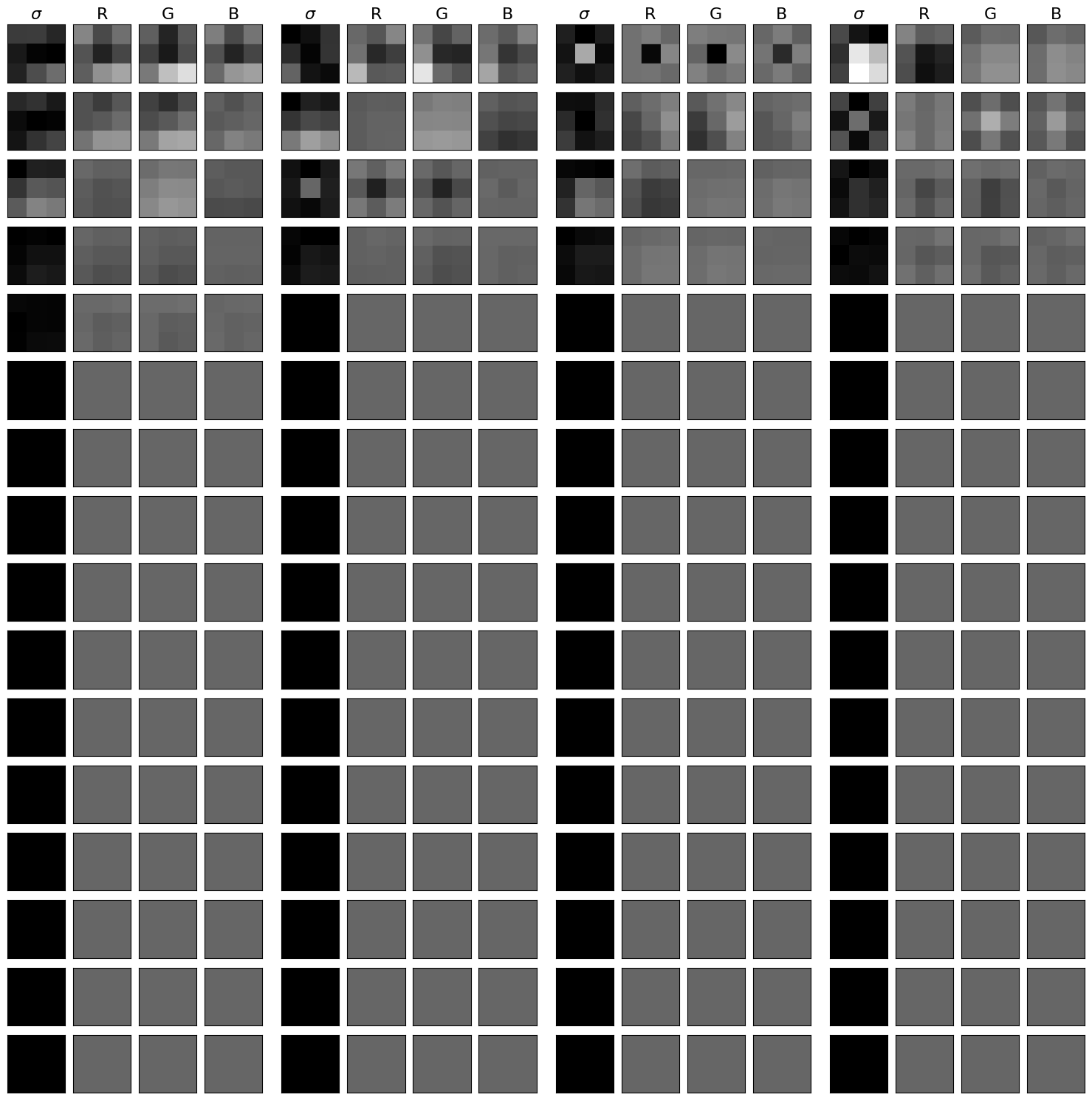}
          \caption{\scriptsize{arcface-r100-glint360k}}\label{adaface}
      \end{subfigure}
  \end{subfigure}
  \caption{\textbf{``Inactiveness" in first convolution filter weight is not specific to any training dataset}. Visualization of 64 convolution filter weight values of the first convolution block of popular pre-trained ArcFace model trained on different training datasets.  Filters are presented in the row-major order based on the L1-norm ranking from top left grid position to the bottom right. The RGB grid uses black for negative weights, gray for zero weights, and white for positive weights. The standard deviation grid uses black for zero standard deviation, gray for half-way standard deviation, and white for positive standard deviation. Notably, a significant portion of filters across all versions trained on different training sets are deemed ``inactive", since their RGB weight grid is predominantly gray, and their standard deviation grid is mostly black. Note that the filters are 
presented in row-major order based on the L1-norm ranking from top left grid position to the bottom right. }
  \label{fig:pretrained_weight_vis}
\end{figure*}

\section{Literature Review}\label{litreview}

Reinitialization of neural network connections holds significance in various tasks, such as training sparse frameworks \cite{han_iclr_2016, zhang_neurocomputing_2022}, and online learning \cite{ash_nips_2020, caccia_clla_2022}, among others. Additionally, it can serve as a valuable regularization approach to improve the network's generalizability \cite{alabdulmohsin_arxiv_2021, zaidi_pmlr_2023, zhou_iclr_2021}. In this section, we provide a brief overview of the relevant works that specifically focus on reinitialization of one or multiple neural network connections during training. 

Reinitialization involves assigning weight values during training that are not the backprop-determined values.
Existing literature explores various approaches to reinitialization. The first and most popular approach is \textit{weight-magnitude initialization} \cite{bellec_iclr_2018, han_iclr_2016, zhang_neurocomputing_2022}. These methods involve reinitializing the weights (or, ``connections'') in neural networks when they are close to zero. In certain studies, the connections or weights are randomly reinitialized to their prior-to-start-of-training values \cite{bellec_iclr_2018, han_iclr_2016}. In contrast, \cite{zhang_neurocomputing_2022} adopts a hierarchical rule for the reinitialization. The second method is referred to as \textit{random-subset initialization} \cite{taha_cvpr_21,ksikazek_iciap_2022}. In this approach, a subset of parameters, of a fixed size, is uniformly chosen at random in each round for reinitialization. The third method is known as \textit{layer-wise reinitialization} \cite{alabdulmohsin_arxiv_2021}. It entails re-scaling a portion of the network to its original norm while reinitializing the remaining parts. Next, we have the fourth method called \textit{fixed-subset reinitialization} \cite{taha_cvpr_21}. In this approach, a subset of parameters is randomly chosen before the training begins and remains fixed throughout the training process. These randomly initialized ``frozen'' parameters serve as re-initialized connections/weights. The fifth method is referred to as \textit{whole-network reinitialization} \cite{ksikazek_iciap_2022}. This method involves reinitializing the entire network's weights if the weight in the bottleneck layer is entirely inactive. The final and probably least common method is \textit{fully-connected-layer reinitialization} \cite{li_icml_2020, zhao_icpr_2018}, %
where either the last fully connected layer or the classifier head of the network is re-initialized.

Previous works look at reinitializing individual weights \cite{han_iclr_2016,zhang_neurocomputing_2022} or some layers of the network \cite{li_icml_2020, zhao_icpr_2018}, whereas our approach is unique in that it re-initializes a block of weights that corresponds to an image filter in the first layer. This is important because the these filters extract features from the input image
\cite{zhang_iclr_2019,zhou_iclr_2021}. We refer to our approach as \textbf{reactivating} the filter, rather than simply reinitializing weights. Previous works simply re-initialize weights during training, using the same random distribution used to assign initial weights in the network before learning starts \cite{han_iclr_2016}. Instead, our approach creates a pattern of weights to reactivate a filter based on the state of the other filters as they exist at that point in the learning process. Before learning has started it can be reasonable to use the same random distribution for all weights, but once learning is in progress it makes better sense to choose weights to reactivate an inactive filter based on the context of what other filters the network has learned to that point. In addition, many previous works reinitialize individual weights at random points in  learning, which can be useful for regularization. Our approach assumes that regularization is provided by other techniques such as data augmentation \cite{perez_arxiv_2017}, drop out \cite{srivastava_jmlr_2014}, batch normalization \cite{ioffe_icml_2015}, weight decay \cite{rumelhart_nature_1986}, early stopping, and so on, and instead uses the event of a filter going inactive as a trigger for reactivation. 

\section{Inactive Filters in First Convolution Layer}\label{inactive_in_pretrained}

\paragraph{Threshold definition.} A convolution filter is categorized as inactive when its presence or exclusion does not affect the network's output. In the context of face recognition using ResNet-based networks of varying depths, the initial convolution layer comprises 64 convolution kernels, each with dimensions of $3 \times 3 \times 3$, resulting in 27 weights per kernel. This kernel functions as a unit and individual weights cannot be treated independently. Therefore, when the substitution of all 27 weights within a single kernel with zeros produces no change in the face embedding vector, that filter is inactive.

The procedure for determining a threshold to identify an inactive filter comprises two steps: a) training a model under standard training setup, and b) zeroing out the weights in the filters within the initial layer if all 27 weights in the kernel fall below the established threshold. We assess the cosine similarity of a randomly selected 10\% of the images in the training set using both a) and b) at various thresholds, and the findings can be summarized as follows:

\begin{equation} \label{theta}
    S_C(A,B) =
    \begin{cases}
        < 1 & \text{if } \theta > 10^{-3} \\
        1 & \text{if } \theta \leq 10^{-3}
    \end{cases}
\end{equation}
where $S_C$ signifies the cosine similarity, $A$ represents the feature embedding obtained from the original model, and $B$ represents the embedding from the model in which kernels are set to zero if all absolute values in the kernel are less than or equal to the specified threshold ($\theta$) used to detect inactive filter.
Put simply, when all of a filter's $3 \times 3 \times 3$ weights in the first  are within
$10^{-3}$ of zero, they can be set to zero and the cosine similarity of the embeddings before and after is 1. 
All results presented in this work uses $\theta = 10^{-3}$ as threshold.

\medskip

\noindent{\bf Inactive Filters in pre-trained ArcFace.} We now highlight the presence of inactive filters in one of a popular open-source face recognition model. As mentioned earlier, the convolution filter in the first layer of ResNet adapted for face recognition is  $3 \times 3 \times 3$ and if setting all weights in a kernel to be zero doesn't change the output, the kernel is effectively inactive. 

Table \ref{tab:threshold-table} displays the count of inactive filters for $\theta=10^{-3}$ and $\theta=10^{-6}$ thresholds, %
revealing two significant observations. One is that a large fraction of the convolution filters in the first layer of widely-used pre-trained face-recognition model are inactive. 
Secondly, the number of inactive filters is generally consistent for both $\theta = 10^{-3}$ and $\theta = 10^{-6}$ threshold values. This provides additional evidence to support the observation mentioned in Equation \ref{theta}, i.e., that using $\theta = 10^{-3}$ is a good choice as a cue for reactivation.

\begin{table}[!h]
\centering
\resizebox{.8\columnwidth}{!}{%
\begin{tabular}{|l|cc|}
\hline
\multicolumn{1}{|c|}{\multirow{2}{*}{Pre-trained-Models}} & \multicolumn{2}{c|}{Number of ``inactive" filters} \\ \cline{2-3} 
\multicolumn{1}{|c|}{} & \multicolumn{1}{c|}{$\theta = 10^{-3}$} & $\theta = 10^{-6}$ \\ \hline
arcface-r100-ms1mv2 & \multicolumn{1}{c|}{38} & 38  \\ \hline
arcface-r18-ms1mv3 & \multicolumn{1}{c|}{31} & 30 \\ \hline
arcface-r34-ms1mv3 & \multicolumn{1}{c|}{47} & 47  \\ \hline
arcface-r50-ms1mv3 & \multicolumn{1}{c|}{44} & 44 \\ \hline
arcface-r100-ms1mv3 & \multicolumn{1}{c|}{43} & 42 \\ \hline
arcface-r18-glint360k & \multicolumn{1}{c|}{43} & 43  \\ \hline
arcface-r34-glint360k & \multicolumn{1}{c|}{45} & 45  \\ \hline
arcface-r50-glint360k & \multicolumn{1}{c|}{45} & 45  \\ \hline
arcface-r100-glint360k & \multicolumn{1}{c|}{47} & 47  \\ \hline
\end{tabular}%
}
\vspace{0.2cm}

    {\small Total Number of Filters ($N$) = 64 \\
    $\theta$ = threshold used to identify `inactive" filters.} 
    
\caption{ \textbf{Number of ``inactive" filters present in first convolution layer of commonly used pre-trained ArcFace face recognition models.} \textit{The presence of an ``inactive" filter is not specific to any particular depth of the ResNet backbone or the training dataset used to train the model}. Up to 73\% of the filters in the initial convolutional layer, out of a total of 64 filters, are considered ``inactive" (as seen in the case of arcface-r100-glint360k). This is a common problem with other face recognition models as well. (See Supplementary Material) }
\label{tab:threshold-table}
\vspace{-0.75em}
\end{table}

\section{Reactivation of Inactive Filters}\label{approaches}
This section explores the motivations, theoretical aspects, and proposed approaches within this work. 

\subsection{Motivations}
The filters learned in the first layer of the network process an input image to extract primitive features that are used by later layers \cite{arpit_icml_2017, zeiler_eccv_2014}.
Hence, an inactive filter signifies a missed opportunity for extracting information from the image. Furthermore, research by Li et al.\cite{li_nips_2018} suggests that increasing the number of active filters may result in a more stable loss landscape, thereby facilitating convergence towards better minima. With this consideration in mind, our objective is to reactivate the inactive filters during training without disturbing the loss landscape, thereby granting the kernels another opportunity to discover a useful filter pattern.

\subsection{Theoretical Considerations}\label{theory}
Consider a convolution kernel (K) applied to a multi-channel input RGB image (V) which produces output (Z). Z then flows through the network to compute the cost function J \footnote{Note: To simplify, subscripts have been removed for channel, image height, and width as the representation shown is for a standard 3D convolution kernel on a 3-channel input image}. Here, we assume Z as the final output on which the loss is computed for simplification and to focus on first convolution layer. The training objective is to minimize the loss function J(V, K). During backpropagation, we receive the tensor G such that,
\begin{equation} \label{tensor_G}
    G = \frac{\delta}{\delta Z} J(V,K)
\end{equation}
To train the network, we need to compute the \textit{derivatives with respect to the weights in the kernel} as:
\begin{equation} \label{gradient_g}
    g(G,V) =  \frac{\delta}{\delta K} J(V,K)
\end{equation}

Equation \ref{gradient_g} reveals that the set of weights in the kernel layer influences the learning trajectory. To train the networks, we apply SGD iteratively to update the parameters of the kernel weights. The second-order approximation of the iterative loss can be further decomposed as \cite{skorski_acml_2021}:
\begin{equation}\label{hessian}
    J(K - \gamma g) \approx J(K) - \gamma \cdot g^T \cdot g + \frac{\gamma^2}{2} \cdot g^T \cdot H \cdot g 
\end{equation}
where, $\cdot$ represents the tensor dot product. In order to guarantee progress in learning, the maximum step size that ensures a decrease is denoted as $\gamma^* = ||H||^{-1}$, as explained in \cite{goodfellow_mitpress_2016} , where $||H||$ represents the Hessian norm. \textit{This value corresponds to the maximal eigenvalue of the Hessian, emphasizing the importance of controlling the norm of Hessian to ensure smooth training with constant step size}. This is especially important in the context of \textit{reactivating weights}, as the introduction of atypical weights could disturb the loss landscape \cite{li_nips_2018}.

\subsection{Approaches}
We have observed that once a filter goes ``inactive,'' it remains inactive in the standard training process, as shown in Figure \ref{fig:weight_track}. This observation is also true for networks with ResNet-backbone for general object recognition \cite{zhang_neurocomputing_2022}. 
Considering the theoretical factors and this observation, we employ reactivation during training through one reasonable method known as directed scaled random reactivation. Drawing on insights and understanding from this strategy, we introduce our most effective technique: \textbf{CRAFT}.

\medskip

\noindent{ \bf Directed scaled random reactivation.}
Perhaps the simplest approach to reinitialize an inactive filter is to give it a new set of weights from the same distribution that was used to assign weights at the start of the learning process. This could potentially introduce chaoticity in the loss landscape \cite{li_nips_2018,skorski_acml_2021}.
Another potential issue with this basic approach is that when filters are reactivated, their weights may have significantly different L1-norm values compared to the active filters, potentially causing the reactivated filter to revert to an inactive state.
\textit{To address the previously mentioned issue, we choose to reactivate the inactive filter weights based on the ranking of the most active filter using the L1-norm criterion}. When an inactive filter is detected, we reactivate the filter by sampling weight values from a Gaussian distribution with the same mean and standard deviation as the filter with the highest L1-norm. This approach is ``directed'' in that it is applied to filters that are detected as inactive, and ``scaled random'' in that the reactivated weight values are from a  Gaussian distribution scaled to the most active filter during training. With this approach, a reactivated filter's position is in proximity to the active filters (with a controlled Hessian norm after reactivation), but it has a controlled level of randomness that allows the kernel weights to be adjusted by backpropagation, thereby potentially facilitating the learning of an additional fundamental feature from input face images.

\medskip

\noindent{\bf CRAFT.} 
Another straightforward method for reactivation could involve hierarchically replacing a detected inactive filter with a currently ``active'' filter. This approach will maintain L1-norm consistency with active filters (also will have controlled Hessian norm after reactivation), all the while utilizing an understanding of the current state of network training.
However, it also exacerbates the widely recognized problem of redundancy in deep neural network learning \cite{casper_aaai_2021, kahatapitiya_wacv_2021}. To address both concerns simultaneously, \textit{we hierarchically replace the inactive filters with the ``complement'' of the current strong active filters.} Using a complement matrix will reverse the direction of the eigenvectors but the magnitude of eigenvalue remains the same. Since the magnitude of the maximal eigenvalue is unchanged, the norm of the Hessian remains controlled after reactivation. This, in turn, does not affect the learning process as shown in Equation \ref{hessian}, \textit{ensuring that the reactivated filter begins at a valid and potentially unexplored point in the filter weight space}. 

The implementation of the method involves evaluating filters at the end of each epoch to determine if any have become inactive. If so, the ``active'' filters are ranked based on importance, and the inactive filters are hierarchically replaced with the complementary of ``active'' counterparts. 
For example, consider the end of epoch 1, where filters 2, 10, and 15 are detected as inactive. After ranking the filters,  say filters 63, 57, and 52 have the largest L1 norms. In a random order, we copy the negative values of filters 63, 57, and 52 to filters 2, 10, and 15, such that the inactive filter is replaced with the complementary of active filters. 

We also consider the scenario where a same filter holds the top L1-norm rankings for multiple epochs and can be employed for reactivation on multiple occasions. Consequently, each active filter is restricted to be used for reactivation only once. In cases where the count of inactive filters exceeds that of the active filters, particularly when all active filters have been utilized as cues for reactivation at least once, any remaining inactive filter randomly selects a complement from the active filters that have already been used. Nonetheless, this situation occurs relatively infrequently during our training.

\section{Implementation Details}\label{implementation}

To implement our reactivation schemes for comparison experiments, we use ArcFace loss  
with ResNet-50 backbone, trained on cleaned WebFace4M dataset \cite{zhu_cvpr_2021,bhatta_arxiv_2023} . For ArcFace, we employ a combined margin setup with margin combination values of (1.0, 0, 0.4). The model is trained for 20 epochs using SGD as the optimizer, with a momentum of 0.9 and an initial learning rate of 0.1. All the mentioned configuration parameters align with the ones utilized for training WebFace4M on the ResNet-50 backbone, as mentioned in insightface \cite{Insightface} GitHub repository. 

We present results for three distinct training techniques: standard training with no reactivation (baseline), directed scaled random reactivation, and CRAFT.
Results for each approach consist of five training runs.
On average, each model requires about 22-24 hours of training on a node equipped with four Titan X GPUs hosted on Intel(R) Xeon(R) CPU E5-2630 v3 @ 2.40GHz. 

\section{Results}\label{results}

\noindent {\bf Evaluation on Standard Benchmarks.}
This section presents the 1:1 verification accuracy ($\%$) for three approaches on the standard benchmark datasets such as LFW \cite{lfw}, CFP-FP \cite{cfpfp}, AGEDB-30 \cite{agedb30}, CPLFW \cite{cplfw}, and CALFW \cite{calfw}. 

It is essential to acknowledge that deep learning training has an inherently random element.
Typically, when introducing new loss functions or training schemes in the literature, only one accuracy number for a single trained model is reported.
This can raise concerns about whether the accuracy numbers are just a fortunate result of a single training run. In this study, we report the mean and standard deviation of accuracy using our proposed technique over 5 independent runs. This enables a more reliable evaluation of the performance.

\setlength\extrarowheight{5pt}
\begin{table}[!h]
\centering
\resizebox{.95\columnwidth}{!}{%
\begin{tabular}{|c|ccc|}
\hline
\begin{tabular}[c]{@{}c@{}}\diagbox{Benchmark Dataset}{Training Scheme}\end{tabular} & \begin{tabular}[c]{@{}c@{}}Baseline \\ Method \end{tabular} & \begin{tabular}[c]{@{}c@{}}Directed \\ Scaled Random  \end{tabular} & \begin{tabular}[c]{@{}c@{}} CRAFT \\ (Ours) \end{tabular} \\ \hline
LFW & \textcolor[HTML]{FD6864}{99.775 $\pm$ 0.032} &   99.803 $\pm$ 0.006  & \textcolor[HTML]{3339FF}{99.810 $\pm$ 0.014}  \\ \hline
CFP-FP & 99.036 $\pm$ 0.064 & \textcolor[HTML]{FD6864} {98.976$\pm$ 0.078} & \textcolor[HTML]{3339FF}{99.060 $\pm$ 0.066} \\ \hline
AGEDB-30 & \textcolor[HTML]{FD6864}{97.560 $\pm$ 0.143} & 97.753 $\pm$ 0.060    & \textcolor[HTML]{3339FF} {97.829 $\pm$ 0.018}  \\ \hline
CALFW & \textcolor[HTML]{FD6864}{95.982 $\pm$ 0.086} &  96.006 $\pm$ 0.016  & \textcolor[HTML]{3339FF} {96.050 $\pm$ 0.062}   \\ \hline
CPLFW & \textcolor[HTML]{FD6864}{94.077 $\pm$ 0.104} & 94.120 $\pm$ 0.010   & \textcolor[HTML]{3339FF} {94.220 $\pm$ 0.003}  \\ \hline \hline
Average Accuracy & \textcolor[HTML]{FD6864} {97.286 $\pm$ 0.085 } &  97.331 $\pm$ 0.034  & \textcolor[HTML]{3339FF} {97.394 $\pm$ 0.032 }  \\ \hline 
\end{tabular}%
}
\caption{ \textbf{Comparing the accuracy of baseline methods with various proposed approaches}. Utilizing reactivation techniques during training consistently produces improved models compared to the standard setup. Particularly, CRAFT outperforms both baseline method and directed scaled random reactivation with the highest average accuracy across all datasets and on average accuracy. Importantly, it's worth highlighting that using any meaningful re-activation method results in a model as good or better than training without reactivation. [Keys: \textcolor[HTML]{3339FF}{\textbf{Best}}, \textcolor[HTML]{FD6864} {\textbf{Worst}}]
}
\label{tab:acc_table}
\end{table}

Table \ref{tab:acc_table} summarizes the accuracy results for the four approaches across the five test sets.
The traditional (no reactivation) training approach has the lowest accuracy on four of the test sets (LFW, AGEDB-30, CALFW, CPLFW).
\textit{This suggests that employing any reasonable reactivation scheme that doesn't introduce chaoticity into the loss landscape can lead to improvements over the conventional deep CNN training pipeline.}

CRAFT achieves the best results, as it obtains the highest accuracy on all five test sets (LFW, CFP-FP, AGEDB-30, CALFW and CPLFW) and the highest average accuracy across the test sets. 
Notably, the average accuracy for AGEDB-30 improves from $97.560$ to $97.829\%$, for CPLFW from $94.077$ to $94.220\%$, for LFW from $99.775$ to $99.810\%$, for CALFW from $95.982$ to $96.050\%$, and for CFP-FP from $99.036\%$ to $96.060\%$. The overall accuracy of model trained with CRAFT increases from $97.286$ to $97.394\%$, which is akin to correctly identifying 65 additional match and non-match pairs of images, on average. 

\noindent {\bf Evaluation on IJBB and IJBC.}
IJBB \cite{white_cvprw_2017} and IJBC \cite{maze_ijcb_2018} are datasets known for being challenging compared to standard face recognition benchmarks. In addition to the results from benchmark dataset, we also present the result of our reactivation scheme applied to the IJB-B dataset and the IJB-C datasets. The 1:1 verification protocol results are summarized in Table \ref{tab:ijb}. Significantly, these findings are consistent across the IJBB and IJBC datasets, where the model trained without reactivation exhibits the worst performance. While scaled reactivation improves accuracy, CRAFT outperforms both methods in terms of performance, increasing the accuracy on IJBB from $95.164\%$ to $95.284\%$ and on IJBC from $96.886\%$ to $96.960\%$.

\setlength\extrarowheight{2pt}
\begin{table}[!h]
\centering
\resizebox{0.8\columnwidth}{!}{%
\begin{tabular}{|c|c|c|}
\hline
\begin{tabular}[c]{@{}c@{}}\diagbox{Training Scheme}{Benchmark Dataset}\end{tabular} & IJB-B & IJB-C \\ \hline
Baseline Method & \textcolor[HTML]{FD6864} {$95.164 \pm 0.052$} & \textcolor[HTML]{FD6864}{$96.886 \pm 0.026$} \\ \hline
Directed Scaled Random & $95.234 \pm 0.076$ & $96.912 \pm 0.017$ \\ \hline
CRAFT (Ours) & \textcolor[HTML]{3339FF}{$95.284 \pm 0.059$}  & \textcolor[HTML]{3339FF}{$96.960 \pm 0.021$}\\ \hline
\end{tabular}%
}
\caption{\textbf{Evaluation on IJB-B and IJB-C: TAR@FAR=0.01\%}. Reactivation enhances the model's performance on more challenging datasets, implying that it helps the network discover important filter patterns with an additional opportunity for learning, a factor overlooked by standard (no-reactivation) training setup. [Keys: \textcolor[HTML]{3339FF}{\textbf{Best}}, \textcolor[HTML]{FD6864} {\textbf{Worst}}]
}
\label{tab:ijb}
\vspace{-1em}
\end{table}

\section{Discussions}
\subsection{How does reactivation increase the accuracy?}

We have demonstrated that the use of reactivation effectively enhances model accuracy. In this section, we will explore the reasons behind this improvement. The boost in accuracy for models trained with reactivation can primarily be attributed to two factors: (a) a reduction in the number of inactive filters and (b) the discovery of additional filter forms, as detailed below.

\vspace{-1em}

\paragraph{Reduced inactive filters} 
As discussed in Section \ref{inactive_in_pretrained}, inactive filters are present in the first layer of the convolution layers.  An inactive filter represents a missed opportunity to capture fundamental information from the input image. Therefore, a reduction in the number of inactive filters in the first layer implies that more information can be made available to subsequent layers. Table \ref{tab:filter_count} presents the average number of inactive filters for baseline training and with the reactivation methods. Both reactivation techniques lead to a reduction in the average number of inactive filters in the final model, resulting in an overall increase in accuracy. \textit{In particular, CRAFT achieves the lowest average number of inactive filters, decreasing it from 28 to 20 – a reduction from 44\% inactive filters to 32\%.} As a result, it delivers the most substantial improvement in accuracy.
\setlength\extrarowheight{2pt}
\begin{table}[!h]
\centering
\resizebox{0.9\columnwidth}{!}{%
\begin{tabular}{|l|c|}
\hline

\multicolumn{1}{|c|}{Training Scheme} & Avg. Number of Inactive Filters $\pm$ std.dev ($\sigma$) \\ \hline
Baseline Method & {\color[HTML]{FD6864} 28 $\pm$ 6}  \\ \hline
Directed Scaled Random &   25 $\pm$ {\color[HTML]{FD6864} 6} \\ \hline
CRAFT (Ours) &   {\color[HTML]{3339FF} 20 $\pm$ 4} \\ \hline
\end{tabular}%
}

\caption{ \textbf{Average Number of ``inactive" filters for different training schemes.}
CRAFT reduces inactive filters on average (lowest mean) and converges to more consistent network (lowest standard deviaton). [Keys: \textcolor[HTML]{3339FF}{\textbf{Best}}, \textcolor[HTML]{FD6864} {\textbf{Worst}}]
 }
\label{tab:filter_count}
\vspace{-1em}
\end{table}

\begin{figure}[!h]
 \centering
  \begin{subfigure}[b]{\columnwidth}
      \begin{subfigure}[b]{0.45\columnwidth}
        \centering
          \includegraphics[width=1\linewidth]{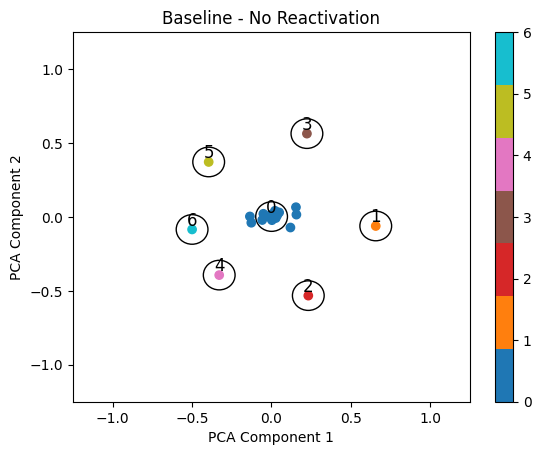}
      \end{subfigure}
      \hfill
      \begin{subfigure}[b]{0.45\columnwidth}
        \centering
          \includegraphics[width=1\linewidth]{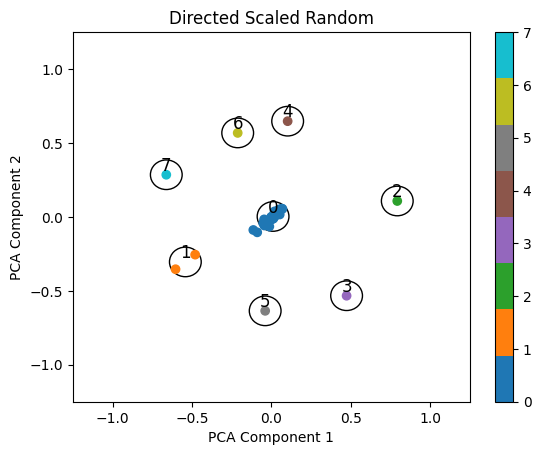}
      \end{subfigure}
      \end{subfigure}
      \begin{subfigure}[b]{\columnwidth}
      \centering
      \begin{subfigure}[b]{\columnwidth}
        \centering
          \includegraphics[width=0.45\linewidth]{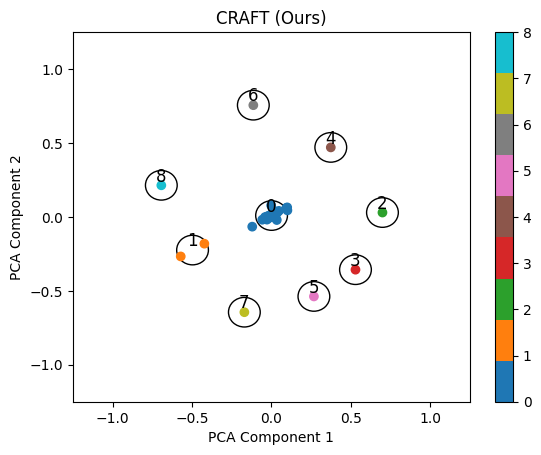}
      \end{subfigure}
  \end{subfigure}
  \caption{\textbf{Mean Shift Clustering on PCA transformed 64 sets $3 \times 3 \times 3$ convolution weights in the first layer for model using baseline training setup and directed complementary reactivation}. 
  Standard Training (no reactivation during training) discovers 7 unique filter types (shown in top left), Training with directed scaled random reactivation discovers 8 unique filter types (shown in top right), and Training with CRAFT discovers 9 unique filter types (shown in bottom). Providing the convolution filter that became ``inactive" during training with a second chance to learn enables the model to identify additional unique filters. }
  \vspace{-1em}
  \label{fig:final_meanshift}
\end{figure}

\paragraph{Increased ``unique'' filter forms} Hubens et al. \cite{hubens_icip_2022} noted the phenomenon of enhanced accuracy resulting from the presence of a diverse range of filters in CNNs. This suggests that the presence of higher number of unique filter-forms that extracts features from images could lead to higher accuracy. In order to check the uniqueness of the filter forms, we perform the mean-shift clustering on PCA transformed $3 \times 3 \times 3$ weights of the first convolution layer of the best model trained without reactivation (baseline) and with reactivation \cite{comaniciu_pami_2002}. From the visualization shown in Figure \ref{fig:final_meanshift}, it is evident that the baseline model trained without any reactivation discovers about 7 unique filter patterns on average, model trained with scaled random reactivation discovers about 8 unique filter patterns on average, whereas \textit{the model trained with CRAFT discovers about 9 unique filter patterns on average}. This indicates that adopting CRAFT during training results in a network with a more robust and varied collection of filter patterns in the first layer, signifying that it has learned additional fundamental filter patterns, thereby enhancing the network's accuracy.

\subsection{Why is CRAFT better?}

Both reactivation methods, directed scaled random reactivation and CRAFT, improve the model's accuracy by reducing the number of inactive filters and discovering additional ``unique" filter forms. Among them, CRAFT stands out as the superior choice. When the reactivation is done using the directed scaled random reactivation, the weights of reactivated filters are in the vicinity of the active filter with some degree of randomness. This reactivation method ensures that the maximal eigenvalue of the Hessian remains under control, as the weights are positioned near the active filter. However, as the reactivated filters start in close proximity to the ``active'' filter, the network is unable to explore a broader search space to optimize their weights during training. Additionally, when viewed from a weight distribution perspective, the filter weights typically deviate from their intialized Gaussian-distributed values after a few training steps. As a result, drawing from a Gaussian distribution, even when scaled to the weights of the active filters, becomes less meaningful (See the distributions in Supplementary Material). As a consequence, this method does not yield a notable reduction in inactive filters and consequently leads to only a marginal improvement in accuracy. 

In the case of CRAFT, when we use complementary filter weights, the maximal eigenvalue has the same magnitude as that of the active filter ensuring controlled norm of Hessian. Now, when we hierarchically employ these complementary filters to reactivate inactive filters, the reactivated filters start at \textit{distinct, valid initial points} in filter weight space. This, in turn, provides the network with an opportunity to explore a broader range of filter weight space during training. Furthermore, from a weight distribution perspective, the complementary reactivation leads to a lateral flip of the filter weight distribution, however it preserves the underlying modality of weight distribution. This, coupled with the fact that reactivated filters begin from a valid, distinct initial point in filter weight space, ultimately leads to an improved model.

\subsection{Are inactive filters universal?}\label{general_vision}

\begin{figure}[!h]
 \centering
      \begin{subfigure}[b]{0.75\columnwidth}
        \centering
          \includegraphics[width=1\linewidth]{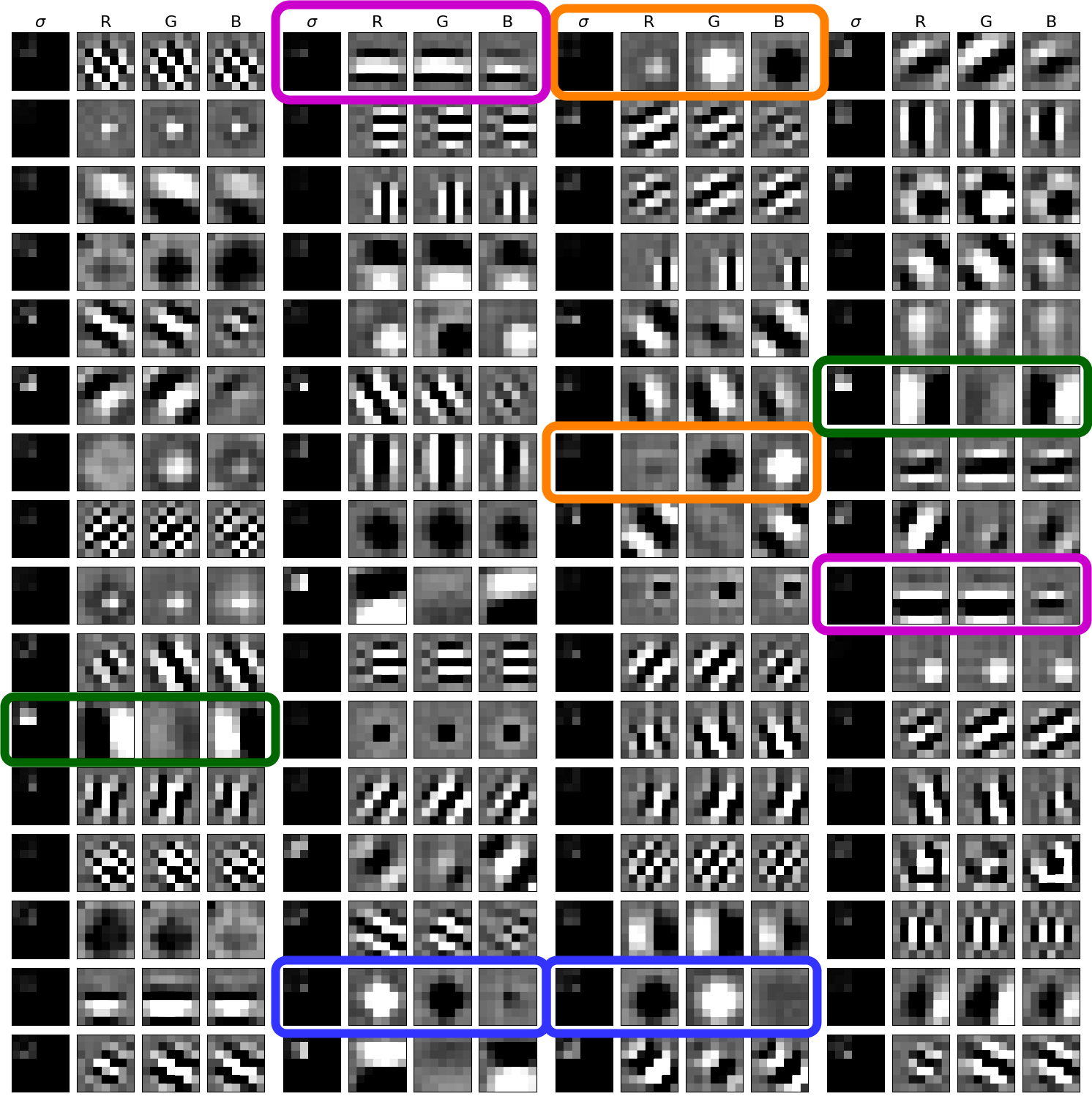}
      \end{subfigure}
  \caption{\textbf{Learned first-layer convolution weights in pre-trained ImageNet}. The figure shows no inactive filters in these pre-trained weights, indicating a reduced likelihood of inactive filters in more general vision tasks. Moreover, the presence of complementary counterparts to learned filters provides additional empirical evidence that using their complementary versions for reactivation might be a sound strategy (complementary filters marked in the same color).}
  \vspace{-0.75em}
  \label{fig:imagenet}
\end{figure}

To investigate the presence of the inactive filter within the broader context of visual tasks, %
we employ the same visualization technique as shown in Figure \ref{fig:pretrained_weight_vis}, but this time, we focus on the pre-trained weights of the Resnet-50 backbone used for ImageNet classification\cite{paszke_nips_2019,deng_cvpr_2009}. Figure \ref{fig:imagenet} illustrates the visualization of the initial layer of the model specifically designed for ImageNet classification. The figure reveals two significant observations: a) The absence of inactive filters implies that as tasks become more general, the likelihood of inactive filters diminishes. b) Models trained for more general vision tasks do exhibit complementary filters (few example complementary filter pairs are highlighted with same color in Figure \ref{fig:imagenet} ). This may serve as additional empirical evidence that the use of complementary active counterparts for reactivation during training is a sensible choice, as shown while adopting CRAFT for face recognition training.

\subsection{Does reactivation always improve accuracy?}
In this section, we compare and contrast the highest and lowest accuracy results obtained through multiple trainings using CRAFT in comparison to the baseline method, which does not use reactivation. From Table \ref{tab:bestvsworstruns}, we observe that the best model achieved using  CRAFT  consistently yields higher accuracy compared to the best model trained with the baseline method, across all five benchmark datasets. Furthermore, the weakest model produced with CRAFT significantly outperforms the weakest model from baseline training, with the exception of CFP-FP, where its performance is only marginally below that of the baseline. This suggests that the model trained with CRAFT consistently converges to a superior model. 

\setlength\extrarowheight{5pt}
\begin{table}[!h]
\centering
\resizebox{0.75\columnwidth}{!}{%
\begin{tabular}{|c|c|c|}
\hline
\multirow{2}{*}{\begin{tabular}[c]{@{}c@{}}\diagbox{Datasets}{Accuracy}\end{tabular}} &
  \begin{tabular}[c]{@{}c@{}}Baseline No-Reactivations\end{tabular} &
  \begin{tabular}[c]{@{}c@{}}CRAFT (Ours)\end{tabular} \\   \cline{2-3}
             & {[}Worst - Best{]} & {[}Worst - Best{]} \\ \hline
LFW          & {\color[HTML]{FD6864}99.71} - 99.81      & 99.78 - { \color[HTML]{3339FF} 99.83}      \\
CFP-FP       & 99.01 - 99.12     & {\color[HTML]{FD6864} 98.96} - {\color[HTML]{3339FF} 99.14}      \\
AGEDB-30     & {\color[HTML]{FD6864} 97.55} - 97.73      &  97.80 - {\color[HTML]{3339FF}97.85}      \\
CALFW        & {\color[HTML]{FD6864} 95.73} - 96.11      & 96.02 - {\color[HTML]{3339FF} 96.12}      \\
CPLFW        & {\color[HTML]{FD6864} 93.90} - 94.18      & 94.18 - {\color[HTML]{3339FF} 94.23}      \\ \hline\hline
Average Acc. & {\color[HTML]{FD6864} 97.18} - 97.39      & 97.34 - {\color[HTML]{3339FF} 97.43}    \\ \hline
\end{tabular}
}
\caption{\textbf{Best and Worst accuracy for baseline training without reactivation and CRAFT over five runs.} Models trained with CRAFT produces at least as good or better model than the baseline method without reactivation. [Keys: \textcolor[HTML]{3339FF}{\textbf{Best}}, \textcolor[HTML]{FD6864} {\textbf{Worst}}] }
\vspace{-1.25em}
\label{tab:bestvsworstruns}
\end{table}

\section{Conclusions}\label{conclusions}

\noindent{\bf Deep CNN face training assumes filter discovery during training is complete.}
As demonstrated in this study and also as discussed in \cite{zhang_neurocomputing_2022},  a convolution filter can become inactive during training.  Conventional CNN training  overlooks this. Allowing filters to go inactive effectively assumes that all filters capable of providing additional useful information have already been identified during training. However, our findings clearly demonstrate that allowing the filters that have gone inactive during training a second chance to learn can lead to  discovery of additional filter patterns and ultimately increase  model accuracy.  

\noindent{\bf Different problems may need different sets of filters.}
Using a deep CNN backbone with 64 filters in the first layer assumes that there exists 64 fundamental filter patterns that the network can learn from images. However, specialized tasks demand fewer ``active'' fundamental features, as observed for face recognition in this work and for biomedical image segmentation networks \cite{mishra_isbi_2019}. In contrast, the general ImageNet classification task appears to involve a higher number of active filters. So, as tasks become more specialized, the frequency of inactive filters increases, highlighting the potential importance of their reactivation.

\noindent{\bf \textcolor[HTML]{FD6864}{Limitations.}}
We show that reactivation of inactive filters improves the current training process for face recognition.  But we do not claim to have found the best possible approach to reactivation, which might vary depending on how the network represents fundamental filters in the early layer.

\section*{Acknowledgement}
The authors would like to thank Dr. Adam Czajka for his invaluable feedback to this work.

{\small
\bibliographystyle{ieee_fullname}
\bibliography{egbib}

\begin{thebibliography}{10}\itemsep=-1pt

\bibitem{Insightface}
Insightface: 2d and 3d face analysis project.
\newblock \url{https://github.com/deepinsight/insightface/}.

\bibitem{alabdulmohsin_arxiv_2021}
Ibrahim Alabdulmohsin, Hartmut Maennel, and Daniel Keysers.
\newblock The impact of reinitialization on generalization in convolutional neural networks.
\newblock {\em arXiv preprint arXiv:2109.00267}, 2021.

\bibitem{arpit_icml_2017}
Devansh Arpit, Stanis{\l}aw Jastrz{\k{e}}bski, Nicolas Ballas, David Krueger, Emmanuel Bengio, Maxinder~S Kanwal, Tegan Maharaj, Asja Fischer, Aaron Courville, Yoshua Bengio, et~al.
\newblock A closer look at memorization in deep networks.
\newblock In {\em Int. Conf. Mach. Learning.}, pages 233--242. PMLR, 2017.

\bibitem{ash_nips_2020}
Jordan~T Ash and Ryan~P Adams.
\newblock On warm-starting neural network training.
\newblock In {\em Adv. Neural Inform. Process. Syst.}, pages 3884--3894, 2020.

\bibitem{bellec_iclr_2018}
Guillaume Bellec, David Kappel, Wolfgang Maass, and Robert Legenstein.
\newblock Deep rewiring: Training very sparse deep networks.
\newblock In {\em Int. Conf. Learn. Represent.}, 2018.

\bibitem{bhatta_arxiv_2023}
Aman Bhatta, Domingo Mery, Haiyu Wu, Joyce Annan, Micheal~C King, and Kevin~W Bowyer.
\newblock Our deep cnn face matchers have developed achromatopsia.
\newblock In {\em arXiv preprint arXiv:2309.05180}, 2023.

\bibitem{caccia_clla_2022}
Lucas Caccia, Jing Xu, Myle Ott, Marcaurelio Ranzato, and Ludovic Denoyer.
\newblock On anytime learning at macroscale.
\newblock In {\em Conf. on Lifelong Learning Agents}, pages 165--182. PMLR, 2022.

\bibitem{casper_aaai_2021}
Stephen Casper, Xavier Boix, Vanessa D'Amario, Ling Guo, Martin Schrimpf, Kasper Vinken, and Gabriel Kreiman.
\newblock Frivolous units: Wider networks are not really that wide.
\newblock In {\em AAAI Conf. Artif. Intell.}, pages 6921--6929, 2021.

\bibitem{comaniciu_pami_2002}
Dorin Comaniciu and Peter Meer.
\newblock Mean shift: A robust approach toward feature space analysis.
\newblock In {\em IEEE Trans. Pattern Anal. Mach. Intell.}, volume~24, pages 603--619. IEEE, 2002.

\bibitem{deng_cvpr_2009}
Jia Deng, Wei Dong, Richard Socher, Li-Jia Li, Kai Li, and Li Fei-Fei.
\newblock Imagenet: A large-scale hierarchical image database.
\newblock In {\em IEEE Conf. Comput. Vis. Pattern Recog.}, pages 248--255. Ieee, 2009.

\bibitem{deng_cvpr_2019}
Jiankang Deng, Jia Guo, Niannan Xue, and Stefanos Zafeiriou.
\newblock Arcface: Additive angular margin loss for deep face recognition.
\newblock In {\em IEEE Conf. Comput. Vis. Pattern Recog.}, pages 4690--4699, 2019.

\bibitem{goodfellow_mitpress_2016}
Ian Goodfellow, Yoshua Bengio, and Aaron Courville.
\newblock {\em Deep learning}.
\newblock MIT press, 2016.

\bibitem{han_iclr_2016}
Song Han, Jeff Pool, Sharan Narang, Huizi Mao, Enhao Gong, Shijian Tang, Erich Elsen, Peter Vajda, Manohar Paluri, John Tran, et~al.
\newblock Dsd: Dense-sparse-dense training for deep neural networks.
\newblock In {\em Int. Conf. Learn. Represent.}, 2016.

\bibitem{he_iccv_2015}
Kaiming He, Xiangyu Zhang, Shaoqing Ren, and Jian Sun.
\newblock Delving deep into rectifiers: Surpassing human-level performance on imagenet classification.
\newblock In {\em Int. Conf. Comput. Vis. Worksh.}, pages 1026--1034, 2015.

\bibitem{lfw}
Gary~B Huang, Marwan Mattar, Tamara Berg, and Eric Learned-Miller.
\newblock Labeled faces in the wild: A database forstudying face recognition in unconstrained environments.
\newblock In {\em Workshop on faces in Real-Life Images: detection, alignment, and recognition}, 2008.

\bibitem{hubens_icip_2022}
Nathan Hubens, Matei Mancas, Bernard Gosselin, Marius Preda, and Titus Zaharia.
\newblock Improve convolutional neural network pruning by maximizing filter variety.
\newblock In {\em IEEE Int. Conf. Image Process.}, pages 379--390. Springer, 2022.

\bibitem{ioffe_icml_2015}
Sergey Ioffe and Christian Szegedy.
\newblock Batch normalization: Accelerating deep network training by reducing internal covariate shift.
\newblock In {\em Int. Conf. Mach. Learning.}, pages 448--456. pmlr, 2015.

\bibitem{kahatapitiya_wacv_2021}
Kumara Kahatapitiya and Ranga Rodrigo.
\newblock Exploiting the redundancy in convolutional filters for parameter reduction.
\newblock In {\em IEEE Conf. Wint. App. Comput. Vis. Worksh.}, pages 1410--1420, 2021.

\bibitem{ksikazek_iciap_2022}
Kamil Ksi{\k{a}}{\.z}ek, Przemys{\l}aw G{\l}omb, Micha{\l} Romaszewski, Micha{\l} Cholewa, Bartosz Grabowski, and Kriszti{\'a}n B{\'u}za.
\newblock Improving autoencoder training performance for hyperspectral unmixing with network reinitialisation.
\newblock In {\em IEEE Int. Conf. Image Process.}, pages 391--403. Springer, 2022.

\bibitem{li_nips_2018}
Hao Li, Zheng Xu, Gavin Taylor, Christoph Studer, and Tom Goldstein.
\newblock Visualizing the loss landscape of neural nets.
\newblock In {\em Adv. Neural Inform. Process. Syst.}, volume~31, 2018.

\bibitem{li_icml_2020}
Xingjian Li, Haoyi Xiong, Haozhe An, Cheng-Zhong Xu, and Dejing Dou.
\newblock Rifle: Backpropagation in depth for deep transfer learning through re-initializing the fully-connected layer.
\newblock In {\em Int. Conf. Mach. Learning.}, pages 6010--6019. PMLR, 2020.

\bibitem{maze_ijcb_2018}
Brianna Maze, Jocelyn Adams, James~A Duncan, Nathan Kalka, Tim Miller, Charles Otto, Anil~K Jain, W~Tyler Niggel, Janet Anderson, Jordan Cheney, et~al.
\newblock Iarpa janus benchmark-c: Face dataset and protocol.
\newblock In {\em Int. Conf. Biometrics}, pages 158--165. IEEE, 2018.

\bibitem{mishra_isbi_2019}
Suraj Mishra, Peixian Liang, Adam Czajka, Danny~Z Chen, and X~Sharon Hu.
\newblock Cc-net: Image complexity guided network compression for biomedical image segmentation.
\newblock In {\em IEEE Int. Symp. Biomed. Img.}, pages 57--60. IEEE, 2019.

\bibitem{agedb30}
Stylianos Moschoglou, Athanasios Papaioannou, Christos Sagonas, Jiankang Deng, Irene Kotsia, and Stefanos Zafeiriou.
\newblock Agedb: the first manually collected, in-the-wild age database.
\newblock In {\em IEEE Conf. Comput. Vis. Pattern Recog. Worksh.}, page~5, 2017.

\bibitem{paszke_nips_2019}
Adam Paszke, Sam Gross, Francisco Massa, Adam Lerer, James Bradbury, Gregory Chanan, Trevor Killeen, Zeming Lin, Natalia Gimelshein, Luca Antiga, et~al.
\newblock Pytorch: An imperative style, high-performance deep learning library.
\newblock {\em Adv. Neural Inform. Process. Syst.}, 32, 2019.

\bibitem{perez_arxiv_2017}
Luis Perez and Jason Wang.
\newblock The effectiveness of data augmentation in image classification using deep learning.
\newblock {\em arXiv preprint arXiv:1712.04621}, 2017.

\bibitem{rumelhart_nature_1986}
David~E Rumelhart, Geoffrey~E Hinton, and Ronald~J Williams.
\newblock Learning representations by back-propagating errors.
\newblock {\em nature}, 323(6088):533--536, 1986.

\bibitem{cfpfp}
S. Sengupta, J.C. Cheng, C.D. Castillo, V.M. Patel, R. Chellappa, and D.W. Jacobs.
\newblock Frontal to profile face verification in the wild.
\newblock In {\em IEEE Conf. Wint. App. Comput. Vis. Worksh.}, February 2016.

\bibitem{skorski_acml_2021}
Maciej Skorski, Alessandro Temperoni, and Martin Theobald.
\newblock Revisiting weight initialization of deep neural networks.
\newblock In {\em Asian. Conf. Mach. Learning.}, pages 1192--1207. PMLR, 2021.

\bibitem{srivastava_jmlr_2014}
Nitish Srivastava, Geoffrey Hinton, Alex Krizhevsky, Ilya Sutskever, and Ruslan Salakhutdinov.
\newblock Dropout: a simple way to prevent neural networks from overfitting.
\newblock {\em The journal of machine learning research}, 15(1):1929--1958, 2014.

\bibitem{taha_cvpr_21}
Ahmed Taha, Abhinav Shrivastava, and Larry~S Davis.
\newblock Knowledge evolution in neural networks.
\newblock In {\em IEEE Conf. Comput. Vis. Pattern Recog.}, pages 12843--12852, 2021.

\bibitem{white_cvprw_2017}
Cameron Whitelam, Emma Taborsky, Austin Blanton, Brianna Maze, Jocelyn Adams, Tim Miller, Nathan Kalka, Anil~K Jain, James~A Duncan, Kristen Allen, et~al.
\newblock Iarpa janus benchmark-b face dataset.
\newblock In {\em IEEE Conf. Comput. Vis. Pattern Recog. Worksh.}, pages 90--98, 2017.

\bibitem{zaidi_pmlr_2023}
Sheheryar Zaidi, Tudor Berariu, Hyunjik Kim, Jorg Bornschein, Claudia Clopath, Yee~Whye Teh, and Razvan Pascanu.
\newblock When does re-initialization work?
\newblock In {\em Adv. Neural Inform. Process. Syst. Worksh.}, pages 12--26. PMLR, 2023.

\bibitem{zeiler_eccv_2014}
Matthew~D Zeiler and Rob Fergus.
\newblock Visualizing and understanding convolutional networks.
\newblock In {\em Eur. Conf. Comput. Vis. Worksh.}, pages 818--833. Springer, 2014.

\bibitem{zhang_iclr_2019}
Chiyuan Zhang, Samy Bengio, Moritz Hardt, Michael~C Mozer, and Yoram Singer.
\newblock Identity crisis: Memorization and generalization under extreme overparameterization.
\newblock In {\em Int. Conf. Learn. Represent.}, 2019.

\bibitem{zhang_neurocomputing_2022}
Ke Zhang, Guangzhe Liu, and Meibo Lv.
\newblock Rufp: Reinitializing unimportant filters for soft pruning.
\newblock {\em Neurocomputing}, 483:311--321, 2022.

\bibitem{zhao_icpr_2018}
Kaikai Zhao, Tetsu Matsukawa, and Einoshin Suzuki.
\newblock Retraining: A simple way to improve the ensemble accuracy of deep neural networks for image classification.
\newblock In {\em Int. Conf. Pattern Recog.}, pages 860--867. IEEE, 2018.

\bibitem{cplfw}
T. Zheng and W. Deng.
\newblock Cross-pose lfw: A database for studying cross-pose face recognition in unconstrained environments.
\newblock Technical Report 18-01, Beijing University of Posts and Telecommunications, February 2018.

\bibitem{calfw}
Tianyue Zheng, Weihong Deng, and Jiani Hu.
\newblock Cross-age lfw: A database for studying cross-age face recognition in unconstrained environments.
\newblock {\em arXiv preprint arXiv:1708.08197}, 2017.

\bibitem{zhou_iclr_2021}
Hattie Zhou, Ankit Vani, Hugo Larochelle, and Aaron Courville.
\newblock Fortuitous forgetting in connectionist networks.
\newblock In {\em Int. Conf. Learn. Represent.}, 2021.

\bibitem{zhu_cvpr_2021}
Zheng Zhu, Guan Huang, Jiankang Deng, Yun Ye, Junjie Huang, Xinze Chen, Jiagang Zhu, Tian Yang, Jiwen Lu, Dalong Du, et~al.
\newblock Webface260m: A benchmark unveiling the power of million-scale deep face recognition.
\newblock In {\em IEEE Conf. Comput. Vis. Pattern Recog.}, pages 10492--10502, 2021.

\end{thebibliography}
}

\end{document}